\documentclass[conference]{IEEEtran}
\IEEEoverridecommandlockouts
\usepackage{cite}
\usepackage{amsmath,amssymb,amsfonts}
\usepackage{algorithmic}
\usepackage{graphicx}
\usepackage{textcomp}
\usepackage{xcolor}
\usepackage[ruled,vlined]{algorithm2e}
\usepackage{caption}
\usepackage{float,subcaption}
\usepackage{tikz}
\usepackage{booktabs}
\usepackage{array}
\usetikzlibrary{shapes,arrows,decorations.pathreplacing}

\tikzstyle{startstop} = [rectangle, rounded corners, minimum width=3cm, text width=4cm, minimum height=1cm, text centered, draw=black, fill=red!30]
\tikzstyle{process} = [rectangle, minimum width=5cm, text width=5cm, minimum height=1cm, text centered, draw=black, fill=orange!30]
\tikzstyle{decision} = [diamond, minimum width=1.5cm, text width=2cm, minimum height=1cm, text centered, draw=black, fill=green!30]
\tikzstyle{arrow} = [thick,->,>=stealth]

\begin{document}

\title{Unsupervised Acute Intracranial Hemorrhage Segmentation With Mixture Models
}

\author{\IEEEauthorblockN{Kimmo Kärkkäinen}
\IEEEauthorblockA{\textit{Computer Science Department} \\
\textit{University of California, Los Angeles}\\
Los Angeles, California, USA \\
kimmo@cs.ucla.edu}
\and
\IEEEauthorblockN{Shayan Fazeli}
\IEEEauthorblockA{\textit{Computer Science Department} \\
\textit{University of California, Los Angeles}\\
Los Angeles, California, USA \\
shayan@cs.ucla.edu}
\and
\IEEEauthorblockN{Majid Sarrafzadeh}
\IEEEauthorblockA{\textit{Computer Science Department} \\
\textit{University of California, Los Angeles}\\
Los Angeles, California, USA \\
majid@cs.ucla.edu}}

\maketitle

\begin{abstract}

Intracranial hemorrhage occurs when blood vessels rupture or leak within the brain tissue or elsewhere inside the skull. It can be caused by physical trauma or by various medical conditions and in many cases leads to death. The treatment must be started as soon as possible, and therefore the hemorrhage should be diagnosed accurately and quickly. The diagnosis is usually performed by a radiologist who analyses a Computed Tomography (CT) scan containing a large number of cross-sectional images throughout the brain. Analysing each image manually can be very time-consuming, but automated techniques can help speed up the process. While much of the recent research has focused on solving this problem by using supervised machine learning algorithms, publicly-available training data remains scarce due to privacy concerns. This problem can be alleviated by unsupervised algorithms. In this paper, we propose a fully-unsupervised algorithm which is based on the mixture models. Our algorithm utilizes the fact that the properties of hemorrhage and healthy tissues follow different distributions, and therefore an appropriate formulation of these distributions allows us to separate them through an Expectation-Maximization process. In addition, our algorithm is able to adaptively determine the number of clusters such that all the hemorrhage regions can be found without including noisy voxels. We demonstrate the results of our algorithm on publicly-available datasets that contain all different hemorrhage types in various sizes and intensities, and our results are compared to earlier unsupervised and supervised algorithms. The results show that our algorithm can outperform the other algorithms with most hemorrhage types.


\end{abstract}


\section{Introduction}



Intracranial hemorrhage is a life-threatening condition that can be caused by either physical trauma or by various medical conditions, such as a high blood pressure or an aneurysm \cite{heit_iv_wintermark_2017}. It is the cause of 15-20\% of strokes, with an estimated 5 million cases per year globally \cite{yousem_nadgir_2017,krishnamurthi2010global}. Depending on the type of the hemorrhage, the mortality rate can be as high as 57\% \cite{haselsberger_pucher_auer_1988}. Approximately half of the deaths occur within the first 48 hours, which is why the treatment must be started as early as possible \cite{broderick1993volume}. Before the treatment can be started, the hemorrhage must be diagnosed from medical images. This process, however, is very time-consuming due to the large number of images per patient and the complexity of the task. Therefore, automating the analysis process allows us to make a diagnosis faster which leads to faster treatment.

Intracranial hemorrhage is typically diagnosed by Computed Tomography (CT) imaging. CT produces a number of cross-sectional slices by combining information from X-ray images taken from different angles. The intensity value of the voxels indicates how much the X-ray was attenuated and it can be used to determine the type of the tissue at each location. Attenuation is measured in Hounsfield Units (HU) which ranges from -1000 for air to 0 for water and to +2000 for dense bones. Brain matter is typically between 20-45 HU while hemorrhage usually starts with a slightly higher attenuation (e.g. 75-85 HU for subdural hemorrhage) but drifts closer to the attenuation levels of the healthy brain matter over time \cite{rao2016dating}. In practice, however, there can be a significant overlap in the intensity values of hemorrhage and healthy tissues. In addition, intracranial hemorrhages can differ in shapes and sizes which further complicates the analysis.

\begin{figure*}
     \centering
     \begin{subfigure}[b]{0.25\textwidth}
         \centering
         \includegraphics[width=\textwidth]{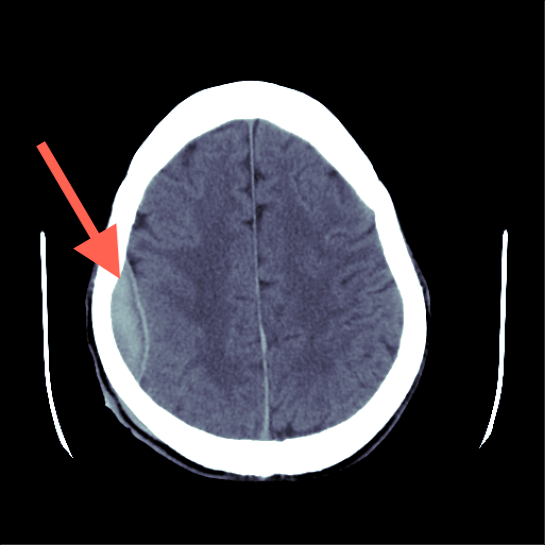}
         \caption{Epidural}
         \label{fig:epidural}
     \end{subfigure}
     \hfill
     \begin{subfigure}[b]{0.25\textwidth}
         \centering
         \includegraphics[width=\textwidth]{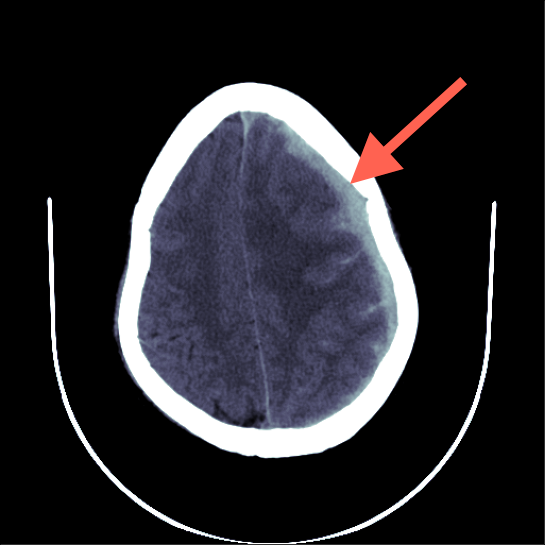}
         \caption{Subdural}
         \label{fig:subdural}
     \end{subfigure}
     \hfill
     \begin{subfigure}[b]{0.25\textwidth}
         \centering
         \includegraphics[width=\textwidth]{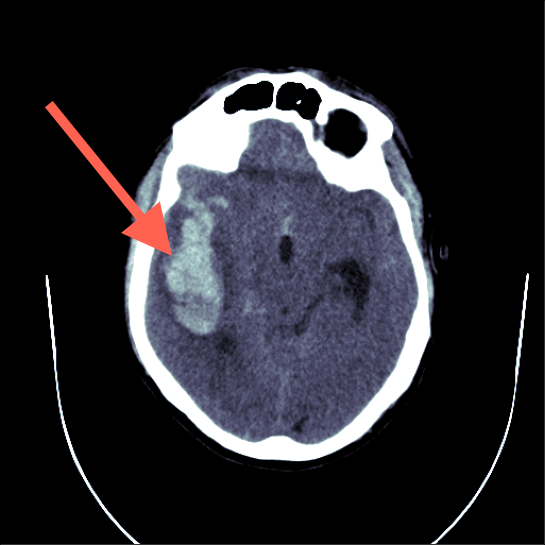}
         \caption{Intraparenchymal}
         \label{fig:intraparenchymal}
     \end{subfigure}
     \hfill
     \par\bigskip
     \hspace*{\fill}
     \begin{subfigure}[b]{0.25\textwidth}
         \centering
         \includegraphics[width=\textwidth]{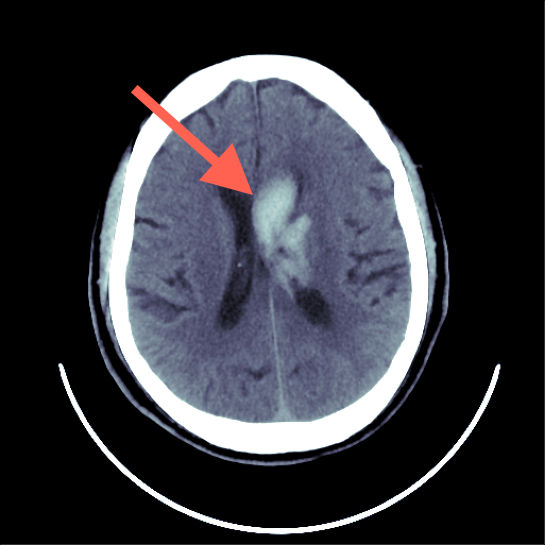}
         \caption{Intraventricular}
         \label{fig:intraventricular}
     \end{subfigure}
     \hfill
     \begin{subfigure}[b]{0.25\textwidth}
         \centering
         \includegraphics[width=\textwidth]{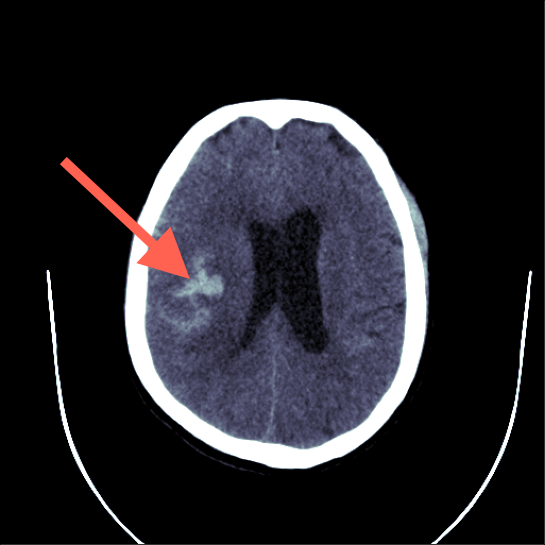}
         \caption{Subarachnoid}
         \label{fig:subarachnoid}
     \end{subfigure}
     \hspace*{\fill}
    \caption{Types of intracranial hemorrhage}
    \label{fig:hemorrhagetypes}
\end{figure*}

There are five categories of hemorrhages with distinct locations and shapes: subdural, epidural, subarachnoid, intraparenchymal, and intraventricular (sample images shown in Figure~\ref{fig:hemorrhagetypes}). One patient can have more than one type of hemorrhage at once. Both epidural and subdural hemorrhage occur close to the skull and they can be distinguished by their shapes: epidural hemorrhage typically has a lentiform shape while subdural hemorrhage has a crescent shape. Intraparenchymal hemorrhage occurs within the brain tissue and typically has a rounded shape. Intraventricular hemorrhage is located inside the brain cavities that are otherwise shown as darker areas close to the center of the brain. Finally, subarachnoid hemorrhage occurs in the space surrounding the brain tissue. While each of these types can have different causes and symptoms, they are typically visible as lighter-than-expected regions in the scan. Other indicators include unusual shapes caused by the pressure, such as a midline that has been pushed slightly in the opposite direction from the hemorrhage.

There have been multiple proposed algorithms for automatic segmentation and classification of CT images (see Section~\ref{section:related}), but much of the recent work is on supervised algorithms which depend on having a large quantity of manually segmented CT scans. As each scan can consist of 30 or more cross-sectional slices (each slice being a two-dimensional image), generating segmented training data is highly time-consuming. In addition, segmentation should be performed by an experienced radiologist due to the complexity of the task. While many research groups have produced such datasets while developing their algorithms, very few of them have been published due to patient privacy concerns, and the few available datasets contain only a small number of scans. Therefore, supervised algorithms can be inaccessible to most unless expert annotators are available. Unsupervised algorithms help alleviate this problem and make the segmentation algorithms more accessible.

In this paper, we propose a fully-unsupervised algorithm for acute intracranial hemorrhage segmentation based on the idea of Mixture Models. We start by demonstrating how to extract the brain from a CT scan containing various tissues and noise. Next, we show how the brain can be represented as a mixture of probability distributions where the location and intensity of different tissue types follow different probability distributions. This representation allows us to find the optimal distribution parameters using the Expectation-Maximization algorithm. We then provide an algorithm for fitting the model when the number of hemorrhage clusters is unknown. Finally, we evaluate our algorithm against other unsupervised and supervised algorithms using publicly-available datasets and provide a visual comparison of the results to demonstrate how the algorithms differ.

\section{Related Works}
\label{section:related}

There are both unsupervised and supervised algorithms for intracranial hemorrhage segmentation. Unsupervised algorithms typically utilize the prior knowledge that hemorrhage voxels tend to have a higher intensity value than healthy voxels. The naive approach would be to determine a hard threshold and mark all voxels above that threshold as hemorrhage and all voxels below that threshold as healthy. In practice, this approach would lead to inaccurate segmentation due to the large overlap of voxel intensities between healthy tissue and hemorrhage. Depending on the chosen threshold, some low-intensity parts of the hemorrhage could be segmented as healthy while some high-intensity parts of healthy tissues could be segmented as hemorrhage. Therefore, intensity threshold alone is not sufficient for segmentation but many algorithms use it as a starting point and refine the results with other techniques.

In earlier works, there have been many suggestions on how to determine the initial intensity threshold. Some techniques include performing Fuzzy C-Means or K-Means clustering on the intensity levels \cite{Bhadauria2013AnImaging,muhammad2017segmentation,bhadauria2014intracranial,gautam2019intracerebral,LONCARIC1995207,sharma2012automatic}, using Otsu's method \cite{al2013automatic,sun2015intracranial}, or comparing the histogram to the expected histogram \cite{RAY2018325}. Once an appropriate threshold has been determined, the results can be refined by using active contouring \cite{Bhadauria2013AnImaging,bhadauria2014intracranial,Liao2010Computer-aidedTomography}, statistical analysis of the clusters \cite{muhammad2017segmentation}, analysis of voxel neighbourhoods \cite{Chuang2006FuzzySegmentation}, or region growing \cite{LONCARIC1995207}.

A different approach proposed by \cite{gillebert2014automated} is to compare the brain scan to a healthy template scan. As each brain scan is unique, the scan must first be transformed into a normalized image. This normalized image is then compared to a template and the differences are scored. Once an abnormal region is detected, the image is transformed back into the original space to be visualized. 

There are also multiple supervised techniques that have been proposed for hemorrhage segmentation. For example, traditional classifiers, such as random forests or logistic regression, have been trained using hand-crafted features on voxels and their neighborhoods \cite{muschelli2017pitchperfect,scherer2016development,alawad2020aibh}. These hand-crafted features have included e.g. voxel intensity, statistics of the voxel's neighbourhood (mean, standard deviation, skew, kurtosis), percentage of high-intensity pixels in the neighbourhood, difference of intensity to the opposite side of the brain, and many more. With the recent advancements on deep learning, there have also been many proposed variants of the U-Net \cite{ronneberger2015unet} architecture \cite{cho2019affinity,li2020deep,arab2020fast,hssayeni2020intracranial,ironside2019fully,sharrock2020intracranial} as well as other neural network architectures \cite{grewal2018radnet,islam2018ichnet}. While supervised algorithms can provide accurate segmentations, they depend on the availability of vast quantities of segmented training data which is rarely published due to privacy concerns.

\section{Methodology}

\subsection{Data}


To develop our algorithm, we used images from a large-scale CT scan dataset published by the Radiological Society of North America (RSNA) as a part of a Kaggle competition \cite{rsna2019intracranial}. The images in this dataset are not segmented but the dataset contains labels for each slice determining the type of the hemorrhage which allows for a visual evaluation of the results. However, the lack of ground-truth segmentations makes it unsuitable for an objective comparison of algorithms.

We therefore used two other publicly-available  datasets for our evaluations. The first dataset has been published by Qure.ai under the Creative Commons license \cite{chilamkurthy2018development}. This dataset consists of 491 CT scans, each scan belonging to a different patient. In addition, each cross-sectional slice has been annotated by a radiologist to denote the type(s) of hemorrhage present in that slice. Out of the 491 patients, 205 patients were labeled as having some type of hemorrhage while the remaining patients were healthy. The mean age of the patients was 48.08 years, and 178 of the patients were female and 313 were male. The scans were performed using six different CT scanner models: GE BrightSpeed, GE Discovery CT750 HD, GE LightSpeed, GE Optima CT660, Philips MX 16-slice, and Philips Access-32 CT. 
The annotations were provided by three senior radiologists and the ground truth label was determined by majority vote. In addition, this dataset has been developed further by \cite{reis2020brain} to include the bounding boxes of hemorrhages within each slice. This bounding box annotation was performed by three trained neuroradiologists. It should be noted that the bounding boxes only indicate where the hemorrhage is located and how large it is approximately, but it does not show us the exact outline of the hemorrhage. Our evaluation will therefore only consider the overlap between our segmention and the bounding box which might contain healthy regions as well.

The second dataset used for evaluation has been published by \cite{hssayeni_2020,hssayeni2020intracranial} and is publicly available on PhysioNet \cite{PhysioNet}. This dataset consists of 82 CT scans with 36 of them containing hemorrhage. Regions with hemorrhage have been delineated by two radiologists who both agreed on the ground-truth segmentations. The mean patient age was 27.8 years with a standard deviation of 19.5. There were 46 male patients and 36 female patients. The CT scanner model was Siemens SOMATOM Definition Edge. As this dataset is very limited in size compared to the Qure.ai dataset but provides accurate outlines of the hemorrhages, we used it only for a visual comparison of the segmentation results.


\begin{figure*}
     \centering
     \begin{subfigure}{0.225\textwidth}
         \centering
         \includegraphics[width=0.99\textwidth]{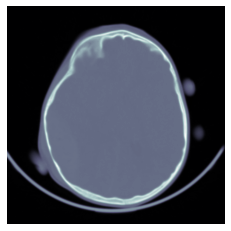}
         \caption{Original image}
         \label{fig:pipeline-orig}
     \end{subfigure}%
     {\LARGE$\xrightarrow{}$}%
     \begin{subfigure}{0.225\textwidth}
         \centering
         \includegraphics[width=0.99\textwidth]{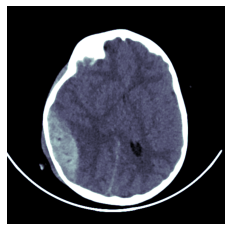}
         \caption{Windowed image}
         \label{fig:pipeline-windowed}
     \end{subfigure}%
     {\LARGE$\xrightarrow{}$}%
     \begin{subfigure}{0.225\textwidth}
         \centering
         \includegraphics[width=0.99\textwidth]{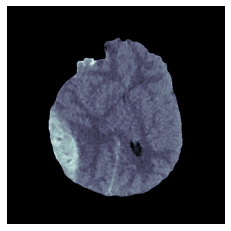}
         \caption{Extracted brain}
         \label{fig:pipeline-brain}
     \end{subfigure}%
     {\LARGE$\xrightarrow{}$}%
     \begin{subfigure}{0.225\textwidth}
         \centering
         \includegraphics[width=0.99\textwidth]{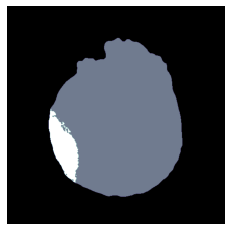}
         \caption{Segmented brain}
         \label{fig:pipeline-segmented}
     \end{subfigure}
    \caption{Image processing pipeline}
    \label{fig:pipeline}
\end{figure*}

\subsection{Preprocessing}

First step of the preprocessing pipeline is to load the CT scan into a 3-dimensional NumPy array where the array dimensions correspond to the x, y, and z coordinates of the voxel and each array entry represents the intensity of that location in Hounsfield Units. An example of one slice of this array is shown in Figure~\ref{fig:pipeline-orig}. Due to the large range of intensity values (air having very low values while bone having very large values), the small intensity variations within the brain are not visible without preprocessing and all brain tissue looks similar. 

Some of the original scans are stored as DICOM files \cite{dicom} while others are stored as NiFTI files \cite{nifti}. Each DICOM file contains only one slice of the scan, so we first load each file for one patient using Pydicom library \cite{pydicom}. These slices are then sorted by using the position metadata and concatenated into a NumPy array. Next, we correct the intensity values by using the slope and intercept values provided in the DICOM metadata using the following linear transformation:

\begin{equation}
    \begin{split}
        \text{intensity} = \text{slope}*v + \text{intercept},
    \end{split}
\end{equation}

where $v$ is the original value. NiFTI files already contain all slices in one file with the correct intensity values, so they are loaded directly into a NumPy array by using NiBabel library \cite{nibabel}.

Next, we perform windowing which removes the very high and very low intensity values which are known to be irrelevant to the segmentation task:

\begin{equation}
    \begin{split}
        \text{intensity}_{\text{windowed}} = \begin{cases}
            i_{\text{min}} &, \text{when intensity} < i_{\text{min}},\\
            i_{\text{max}} &, \text{when intensity} > i_{\text{max}},\\
            \text{intensity} &, \text{otherwise}
        \end{cases}
    \end{split}
\end{equation}

where $i_{\text{min}}$ and $i_{\text{max}}$ are the window boundaries. The boundaries should be chosen s.t. we remove as much of the irrelevant variations as possible without losing any information related to the brain tissue and hemorrhage. In this paper, we have chosen conservative limits of $i_{\text{min}}=0$ and $i_{\text{max}}=100$ which is a wide enough range to cover all of the brain matter and blood while at the same time ignoring the intensity variations of the other tissues, such as bones. It also makes visual evaluation of the images feasible, as we can focus on the minor differences in this narrow range. An example of a windowed image is shown in Figure~\ref{fig:pipeline-windowed}.

After windowing, all of the bone tissue has the same intensity, which makes it easier to remove. We start by selecting all tissues with intensity $i_{\text{max}}$ as the removal mask. This mask might contain gaps due to possible skull fractures, so we fill the small gaps by using morphological closing. The areas covered by this mask are then set to $i_{\text{min}}$. After removing the skull, there are still some soft tissues which are not part of the brain. We remove them by finding the largest contiguous region which we assume to be the brain. We set all regions that are not connected to the largest contiguous region to $i_{\text{min}}$. In addition, a narrow slice along the edge of the brain is removed by using morphological erosion, as it often has a higher intensity and might be mistakenly detected as hemorrhage. An example of a fully preprocessed image can be seen in Figure~\ref{fig:pipeline-brain}.

\subsection{Mixture Model}

After all non-brain voxels have been removed from the scan, our goal is to determine which voxels correspond to healthy tissues and which voxels correspond to hemorrhage. Our model makes the assumption that the different tissue types follow different distributions and therefore we can represent the brain scan using a mixture model. This mixture model should take into account both the location and the intensity of the voxel. For example, a high intensity value could be caused by either hemorrhage or noise depending on whether it is surrounded by a larger region of high-intensity voxels or not. We also make the assumption that the hemorrhage voxels follow a Gaussian distribution in both the coordinate and the intensity space, i.e. the hemorrhage voxels are located close together and have similar intensity values. However, the healthy voxels are spread evenly throughout the brain, so we assume that their intensity values follow a Gaussian distribution but the location follows a Uniform distribution. This approach differs from the earlier algorithms that start by considering only the intensity values and then refine the results by taking coordinates into account as a secondary step. Our algorithm is able to optimize the results in both spaces simultaneously.

Similar intensity values typically represent similar tissue types regardless of the location, so we start with the assumption that the voxel intensity is independent of the voxel location:

\begin{equation}
    P(int,x,y,z | c) = P(int | c) P(x,y,z | c)
\end{equation}

Here, $int$ is the intensity of a voxel in coordinates $(x,y,z)$ and $c$ is the cluster index. For convenience, we define that the first cluster always corresponds to the healthy tissue and there might be additional clusters that correspond to hemorrhage. We also make the assumption that the intensity of each cluster $c$ follows a Gaussian distribution:

\begin{equation}
\begin{split}
    P(int | c) &\sim \mathcal{N}(\mu_{c}, \sigma_{c}), 
\end{split}
\end{equation}

where $\mu_{c}$ and $\sigma_{c}$ are the mean and standard deviation of the intensity values of cluster $c$ respectively. If there are multiple hemorrhage clusters, this allows them to have different intensity distributions, which is necessary because the intensity values could differ based on the type and age of the hemorrhage. As mentioned earlier, we make the assumption that the location of healthy voxels follows a uniform distribution while the location of hemorrhage voxels follows a Gaussian distribution:

\begin{equation}
\begin{split}
    P(x,y,z | c) &\sim \begin{cases}
        \mathcal{U}(0, N_{\text{bv}}) &, \text{when $c=0$}\\
        \mathcal{N}(\mathbf{\mu_{loc,c}}, \mathbf{\Sigma}_{\text{loc,c}}) &, \text{when $c \geq 1$}
    \end{cases} 
\end{split}
\end{equation}

$N_{bv}$ is the number of brain voxels (excluding skull and outside areas), $\mathbf{\mu_{loc,c}}$ is the center of cluster $c$, and $\mathbf{\Sigma}_{\text{loc,c}}$ is the covariance matrix of the location. To determine the optimal parameter values, we use the Expectation Maximization (EM) algorithm \cite{dempster1977maximum}, which is an iterative approach for finding the missing parameters by alternating between calculating the expected cluster membership values (E-step) and calculating the missing parameters (M-step). Using the previously defined distributions, the update rule for the cluster memberships (E-step) becomes:

\begin{equation}
\begin{split}
    \gamma_{i,c} &= \frac{\pi_c p(i|c)}{p(i)} \\
                 &= 
                 \begin{cases}
                 \frac{\pi_c \mathcal{N}(\mu_{c}, \sigma_{c}) \mathcal{N}(\mathbf{\mu_{loc,c}}, \mathbf{\Sigma_{\text{loc,c}}})}{p(int,x,y,z)} & \text{, when $c=0$}, \\
                 \frac{\pi_c \mathcal{N}(\mu_{int,c}, \sigma_{int,c}) \mathcal{U}(0, n_{\text{bv}})}{p(int,x,y,z)} & \text{, otherwise}
                 \end{cases}
\end{split}
\end{equation}

For simplicity, we represent voxels using one-dimensional indices $i$. Here, $\gamma_{i,c}$ is the probability of voxel $i$ belonging in cluster $c$, and $\pi_c$ is the probability of cluster $c$. The cluster memberships are soft, i.e. each voxel can be a partial member of multiple clusters during the optimization process. We can then derive the update rules for the unknown parameters as follows (M-step):

\begin{equation}
\begin{split}
    N_c &= \sum_{i=0}^{N} \gamma_{i,c} \\
    \pi_c &= \frac{N_c}{N} \\
    \boldsymbol{\mu}_{loc,c} &= \frac{1}{N_c} \sum_{i=0}^{N} \gamma_{i,c} \mathbf{coord}_i  \\
    \boldsymbol{\Sigma}_{\text{loc,c}}^2 &= \frac{1}{N_c} \sum_{i=0}^{N} \gamma_{i,c} (\mathbf{coord}_i-\boldsymbol{\mu}_{loc,c})(\mathbf{coord}_i-\boldsymbol{\mu}_{loc,c})^T \\
    \mu_{int,c} &= \frac{1}{N_c} \sum_{i=0}^{N} \gamma_{i,c} int_i \\
    \sigma_{\text{int,c}}^2 &= \frac{1}{N_c} \sum_{i=0}^{N} \gamma_{i,c} (int_i-\mu_{int,c})(int_i-\mu_{int,c})^T
\end{split}
\end{equation}

, where $N_c$ is the effective number of elements in cluster $c$, $\pi_c$ is the proportion of voxels belonging to cluster $c$, $\mathbf{coord_i}$ is a vector containing the coordinates for voxel $i$, and $int_i$ is the intensity of voxel $i$.

Traditional EM algorithm iterates between the E- and M-steps until the solution converges. However, the algorithm expects that the number of clusters is known ahead of time, which is not the case with previously unseen CT scans. Therefore, we need to add an additional step to determine if there are any hemorrhage clusters in the beginning and if we should add more clusters to better represent the hemorrhage regions once the EM algorithm has converged.

First, we can use prior knowledge of the problem domain to determine which voxels we have a high certainty about. Typically, brain tissue has lower intensity values, so we start by initializing all voxels with intensity values below 40 to belong in the healthy cluster. Determining which voxels contain hemorrhage is more challenging because high intensity values could be either hemorrhage or noise. Therefore, we look for large contiguous regions with high intensity values ($>50~\text{HU}$). If one exists, we create a new cluster containing these voxels. This allows us to calculate the initial cluster statistics and start optimizing the clusters by using the EM algorithm. After optimizing these clusters, we look for any remaining large high-intensity regions that do not belong to hemorrhage clusters and create new hemorrhage clusters for them. This process is repeated until we have no remaining high-intensity regions that do not belong in hemorrhage clusters. As a result of this approach, we can have a contiguous hemorrhage region that is represented by multiple clusters if the shape cannot be represented by a single Gaussian distribution (e.g. V shape). Therefore, the number of clusters does not necessarily correspond to the number of distinct hemorrhage regions.

\subsection{Post-processing}

After the EM algorithm has finished, we have a soft clustering where each voxel can belong in multiple different clusters with certain probabilities and there might be multiple hemorrhage clusters. Our goal, however, is to mark each voxel either as healthy or as hemorrhage. To do so, we take the sum of the probabilities of all hemorrhage clusters and compare it to the probability of the healthy cluster. If the combined probability of the hemorrhage clusters is higher than the probability of the healthy cluster, we mark the voxel as hemorrhage:

\begin{equation}
\begin{split}
    \text{label}_{i} = I(\sum_{c \geq 1} p(c|i) > p(c=0|i)), 
\end{split}
\end{equation}

where $I$ is the indicator function. Typically, post-processing is also needed to remove noise from the segmentation results. However, as the optimization process already takes into account the pixel neighbourhoods, i.e. a high-intensity voxel is only marked as hemorrhage if it is located near other high-intensity voxels, it is unlikely to mark noisy voxels as hemorrhage. If a noisy voxel is located near hemorrhage, it would be difficult to distinguish from actual hemorrhage even for a human annotator. Therefore, we did not find including noise removal beneficial with our algorithm. However, in some cases, the hemorrhage regions contain low-intensity holes, which we fill by performing morphological closing. This will fill the small holes while leaving the larger gaps intact which corresponds with how a radiologist would outline the hemorrhage.

\begin{figure*}
\centering
\resizebox{\columnwidth}{!}{%
    \begin{tikzpicture}[node distance=2cm, auto]
    \node (start) [startstop] {Start};
    \node (createnumpy) [process, below of=start] {Convert DICOM/NIfTI to NumPy array};
    \node (windowing) [process, below of=createnumpy] {Perform windowing};
    \node (skullextract) [process, below of=windowing] {
    Remove skull and other non-brain tissues};
    \node (findlargest) [process, below of=skullextract] {Find largest high-intensity region};
    \node (largeenough) [decision, below of=findlargest, yshift=-1cm] {Larger than threshold?};
    \node (nolarge) [startstop, right of=largeenough, xshift=4cm] {Finish, no hemorrhage found};
    \node (calcstats) [process, below of=largeenough, yshift=-1cm] {Update cluster statistics};
    \node (optimizeem) [process, below of=calcstats] {Perform Expectation-Maximization};
    \node (findnew) [process, below of=optimizeem] {Find large high-intensity regions outside existing hemorrhage cluster(s)};
    \node (didfind) [decision, below of=findnew, yshift=-1cm] {Found new regions?};
    \node (addclusters) [process, right of=didfind, xshift=4cm] {Add new cluster(s)};
    \node (probmask) [process, below of=didfind, yshift=-1cm] {Convert probabilities to hard labels};
    \node (filling) [process, below of=probmask] {Fill small holes with morphological closing};
    \node (notfound) [startstop, below of=filling] {Finish};
    
    \draw [arrow] (start) -- (createnumpy);
    \draw [arrow] (createnumpy) -- (windowing);
    \draw [arrow] (windowing) -- (skullextract);
    \draw [arrow] (skullextract) -- (findlargest);
    \draw [arrow] (findlargest) -- (largeenough);
    \draw [arrow] (largeenough) -- node {No} (nolarge);
    \draw [arrow] (largeenough) -- node {Yes} (calcstats);
    \draw [arrow] (calcstats) -- (optimizeem);
    \draw [arrow] (optimizeem) -- (findnew);
    \draw [arrow] (findnew) -- (didfind);
    \draw [arrow] (didfind) -- node[anchor=west] {No} (probmask);
    \draw [arrow] (didfind) -- node[anchor=south] {Yes} (addclusters);
    \draw [arrow] (probmask) -- (filling);
    \draw [arrow] (filling) -- (notfound);
    \draw [arrow] (addclusters) |- (calcstats);
    
    \draw [decorate,decoration={brace,amplitude=16pt,mirror,raise=10pt},yshift=0pt] (createnumpy.north -| createnumpy.west) -- (skullextract.south -| skullextract.west) node [black,rotate=90,midway,xshift=-1.45cm,yshift=1.5cm] {\Large Preprocessing};
    \draw [decorate,decoration={brace,amplitude=16pt,mirror,raise=10pt},yshift=0pt] (findlargest.north -| createnumpy.west) -- (didfind.south -| createnumpy.west) node [black,rotate=90,midway,xshift=-1.35cm,yshift=1.5cm] {\Large Optimization};
    \draw [decorate,decoration={brace,amplitude=16pt,mirror,raise=10pt},yshift=0pt] (probmask.north -| createnumpy.west) -- (filling.south -| createnumpy.west) node [black,rotate=90,midway,xshift=-1.6cm,yshift=1.5cm] {\Large Postprocessing};
\end{tikzpicture}%
}
    \caption{Hemorrhage segmentation process}
    \label{alg:gmm}
\end{figure*}
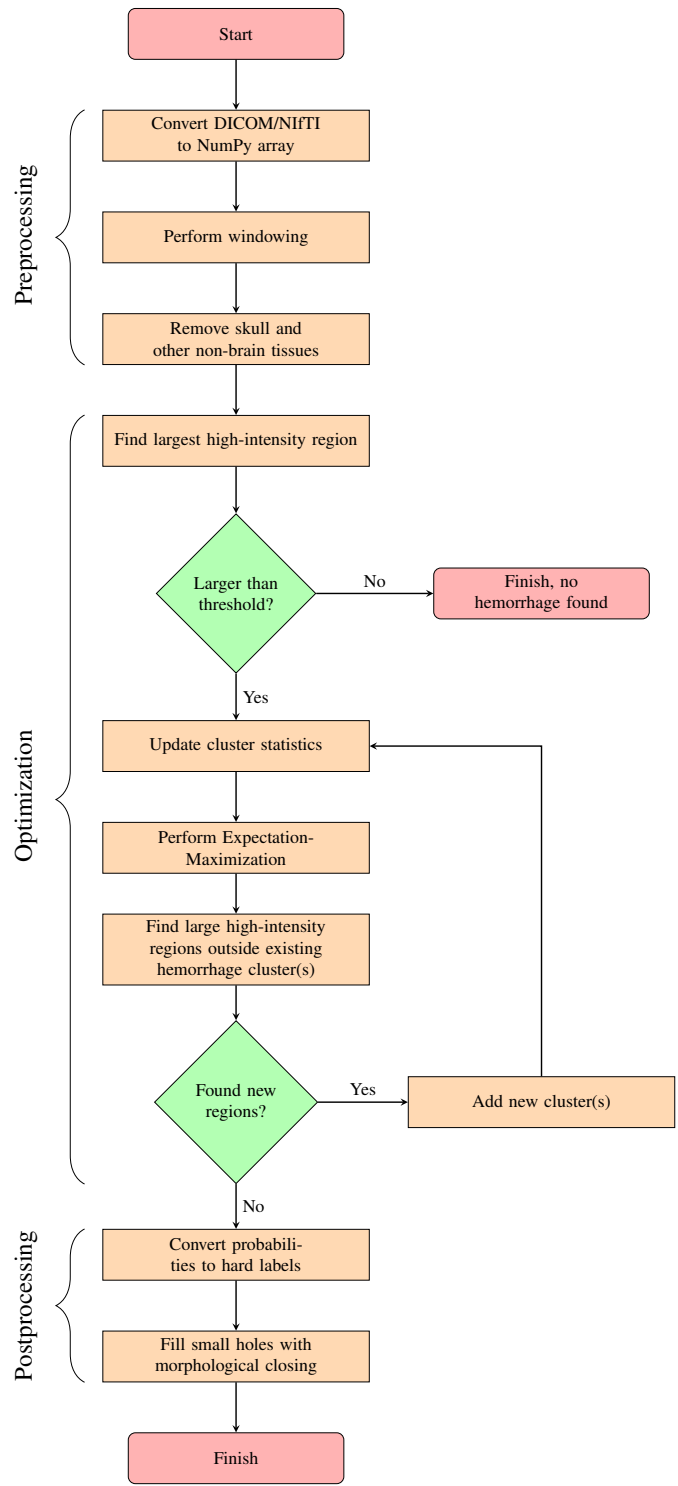

Our entire algorithm is outlined in Figure~\ref{alg:gmm}.


\section{Experiments And Results}

\subsection{Quantitative Evaluation}

We compare our algorithm to Fuzzy C-Means (FCM40, FCM45), PItcHPERFeCT \cite{muschelli2017pitchperfect}, and DeepBleed \cite{sharrock2020intracranial}. FCM is an unsupervised algorithm that finds clusters based on the voxel intensity. It fits multiple Gaussian distributions on the intensity distribution and marks the clusters with a high mean intensity as hemorrhage. For the purposes of our comparisons, we provide results for FCM using two intensity thresholds: 40 and 45. The first threshold was chosen because it achieved the highest Dice scores in our experiments while the second threshold was chosen because it is less likely to include noise in the results. Our implementation of FCM uses the same preprocessing pipeline as our proposed algorithm and its results are postprocessed with morphological opening to eliminate noisy voxels. PItcHPERFeCT and DeepBleed are supervised models that have an open source implementation available, and we use the pretrained models published by the original authors. 

\begin{table*}
\centering
\caption{Dice scores on CQ500 dataset}
\label{table:dice}
\resizebox{\hsize}{!}{%
 \begin{tabular}{l | c c c | c c c | c c c | c c c | c c c}
 \toprule
  & \multicolumn{3}{c|}{Our Model} & \multicolumn{3}{c|}{FCM40} & \multicolumn{3}{c|}{FCM45} & \multicolumn{3}{c|}{PItcHPERFeCT} & \multicolumn{3}{c}{DeepBleed} \\
  Hemorrhage type & Mean & Max & Std & Mean & Max & Std & Mean & Max & Std & Mean & Max & Std & Mean & Max & Std \\ \midrule
 
 All patients (N=196) & \textbf{.197} & \textbf{.814} & .222 & .149 & .563 & .137 & .110 & .504 & .129 & .090 & .556 & .143 & .141 & .684 & .184 \\
 Patients w/ intraparenchymal (N=137) & \textbf{.237} & \textbf{.814} & .218 & .162 & .476 & .138 & .128 & .504 & .136 & .123 & .556 & .162 & .207 & .684 & .200 \\
 Patients w/ subdural (N=64) & \textbf{.244} & \textbf{.733} & .241 & .190 & .563 & .130 & .139 & .406 & .119 & .056 & .345 & .081 & .066 & .402 & .101 \\
 Patients w/ epidural (N=6) & \textbf{.245} & \textbf{.499} & .180 & .191 & .364 & .010 & .176 & .364 & .128 & .118 & .342 & .123 & .142 & .402 & .145 \\
 Patients w/ subarachnoid (N=113) & \textbf{.221} & \textbf{.814} & .233 & .175 & .563 & .140 & .119 & .432 & .123 & .074 & .556 & .113 & .092 & .563 & .129 \\
 Patients w/ intraventricular (N=40) & \textbf{.363} & \textbf{.814} & .229 & .272 & .465 & .129 & .213 & .504 & .136 & .170 & .556 & .175 & .194 & .594 & .192 \\
 Patients w/ only intraparenchymal (N=45) & .128 & .554 & .172 & .059 & .386 & .082 & .045 & .399 & .096 & .108 & .507 & .169 & \textbf{.265} & \textbf{.684} & .226 \\
 Patients w/ only subdural (N=22) & \textbf{.197} & \textbf{.733} & .256 & .177 & .496 & .127 & .139 & .406 & .128 & .040 & .222 & .063 & .035 & .189 & .061 \\
 Patients w/ only epidural (N=1) & \textbf{.027} & \textbf{.027} & .000 & .019 & .019 & .000 & .032 & .032 & .000 & .000 & .000 & .000 & .000 & .000 & .000 \\
 Patients w/ only subarachnoid (N=28) & \textbf{.022} & \textbf{.275} & .065 & .056 & .244 & .067 & .017 & .169 & .041 & .014 & .163 & .041 & .012 & .244 & 067 \\
 Patients w/ only intraventricular (N=0) & - & - & - & - & - & - & - & - & - & - & - & - & - & - & - \\
  \bottomrule
 \end{tabular}%
 }
\end{table*}

We first evaluate these algorithms using the CQ500 dataset. The dataset contains 205 hemorrhage patients but we chose to exclude 9 of them who had chronic hemorrhage. Acute and chronic hemorrhages manifest differently and therefore need different approaches, and none of the algorithms in our comparison have been designed for detecting chronic hemorrhages. As a result, we had 196 hemorrhage patients in our comparison for whom we had bounding boxes delineating the hemorrhage regions in each cross-sectional slice. To evaluate the correctness of the segmentation, we calculate the Dice score which determines the level of similarity between two segmentations and is defined as: 

\begin{equation}
    \text{Dice} = \frac{2*|X\cap Y|}{|X|+|Y|},
\end{equation}

where $X$ and $Y$ are sets of voxels that have been marked as hemorrhage. While Dice score typically ranges from 0 (no matching voxels) to 1 (perfectly matching voxels), our experiments compare segmented images to bounding boxes which typically include healthy tissue as well. As a result, the score of a perfect segmentation would be lower than 1 in practice.

The results can be found in Table~\ref{table:dice}. We first show the results using all of the patients and then by dividing the patients into subgroups based on their hemorrhage types. As one patient can have one or more types of hemorrhage, we further divide the subgroups into patients who have a certain hemorrhage type and possibly other types as well, and into patients who have only a certain hemorrhage type. 

As the results show, our model reaches higher mean and maximum scores than the other models in all but one hemorrhage type. In the case of intraparenchymal hemorrhage, DeepBleed is able to segment the hemorrhages more accurately than our model, but our model still outperforms FCM40, FCM45, and PItcHPERFeCT models. The most significant differences between our model and the supervised models occur with the subdural and subarachnoid hemorrhage types.

\begin{figure*}
     \centering
     \begin{subfigure}[b]{0.49\textwidth}
         \centering
         \includegraphics[width=\textwidth]{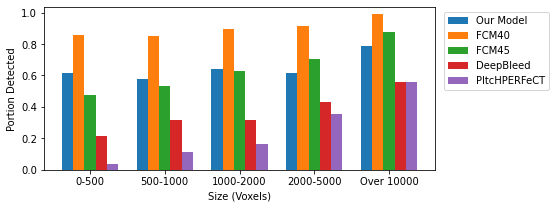}
         \caption{Bounding box size}
         \label{fig:result-size}
     \end{subfigure}
     \hfill
     \begin{subfigure}[b]{0.49\textwidth}
         \centering
         \includegraphics[width=\textwidth]{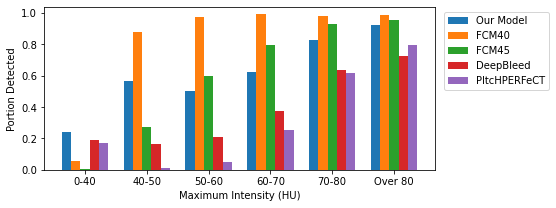}
         \caption{Maximum intensity}
         \label{fig:result-max}
     \end{subfigure}
     \hfill
     \par\bigskip
     \begin{subfigure}[b]{0.49\textwidth}
         \centering
         \includegraphics[width=\textwidth]{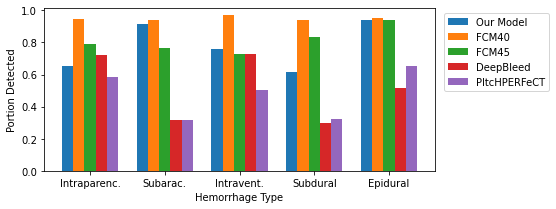}
         \caption{Hemorrhage type}
         \label{fig:result-type}
     \end{subfigure}
     
    \caption{Comparison of hemorrhage detection rates}
    \label{fig:detectionrates}
\end{figure*}

To further understand the situations where the models provide differing results, we inspect the detection rate of bounding boxes based on the hemorrhage characteristics. First, we compare the results on different bounding box sizes (see Figure~\ref{fig:result-size}). The results show that all models perform better with larger hemorrhage sizes, which is as expected, as the smaller hemorrhages are more difficult to distinguish from noise. The size has a large effect especially on PItcHPERFeCT (increase from 4~\% to 56~\%) and DeepBleed (increase from 21~\% to 56~\%), while FCM models (86~\% to 99~\%, 48~\% to 88~\%) as well as our model (62~\% to 78~\%) are slightly less affected. While FCM40 provides the highest detection rate with all hemorrhage sizes, it should be noted that it is more likely to include noise as well, which is evidenced by the lower Dice scores.

When comparing the results on different maximum intensities (see Figure~\ref{fig:result-max}), we can again see that all the algorithms perform worse on low-intensity hemorrhages as expected. The most common situation where the intensity is low is when the hemorrhage is older, but in some less-frequent situations even acute hemorrhages can have a low intensity. The detection becomes more challenging in these cases, as the intensity is similar to healthy brain tissue's intensity, and therefore a radiologist might have to look for other signs of hemorrhage. As FCM40 and FCM45 are focused on the voxel intensities, their detection rate increases rapidly as the intensity level increases. Our model's detection rate increases very early as well, while DeepBleed and PItcHPERFeCT reach high detection rates only when the voxel intensities are over 70~HU.

Finally, when looking at the hemorrhage types (see Figure~\ref{fig:result-type}), we can see that the FCM models provide similar results on each type. This is to be expected, as they only look at voxel intensities without taking the shape or location of the hemorrhage into account. PItcHPERFeCT and DeepBleed show lower detection rates on subarachnoid and subdural hemorrhages, which matches with our observations on the lower Dice scores as well. Our model shows small variations on the detection rates with the highest detection rates occurring on subarachnoid and epidural hemorrhages.

\subsection{Visual Evaluation}

\begin{figure*}
     \centering
     \begin{subfigure}[b]{1.0\textwidth}
         \centering
         \includegraphics[width=\textwidth]{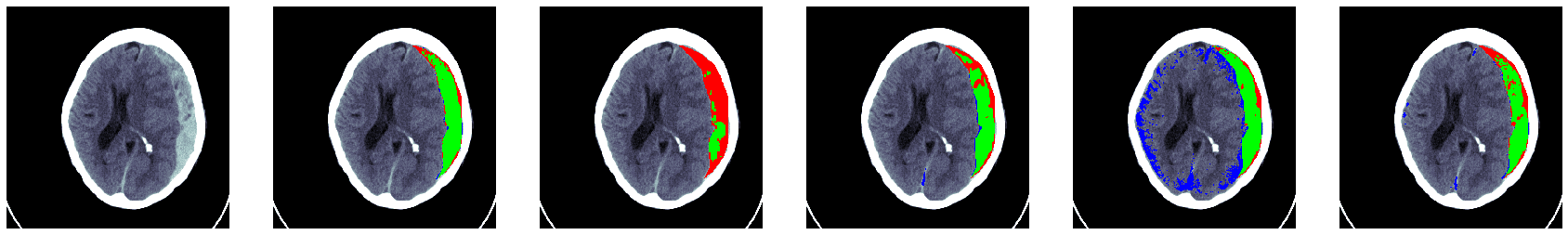}
         \caption{}
         \label{fig:visual1}
     \end{subfigure}
     \par\medskip
     \begin{subfigure}[b]{1.0\textwidth}
         \centering
         \includegraphics[width=\textwidth]{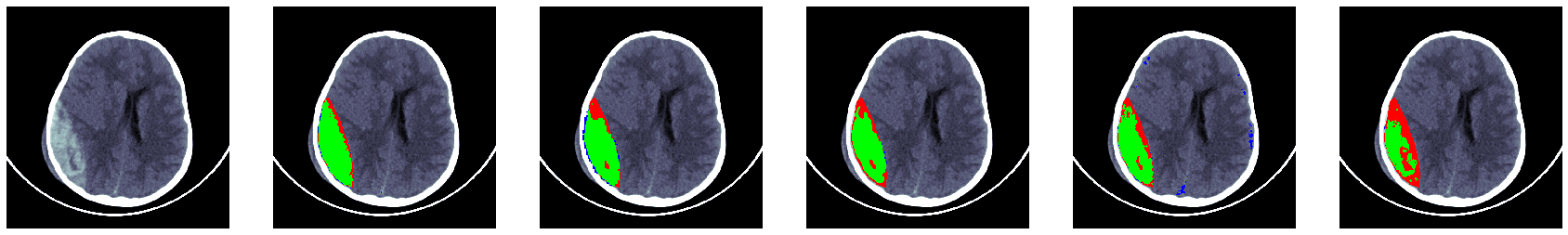}
         \caption{}
         \label{fig:visual2}
     \end{subfigure}
     \par\medskip
     \begin{subfigure}[b]{1.0\textwidth}
         \centering
         \includegraphics[width=\textwidth]{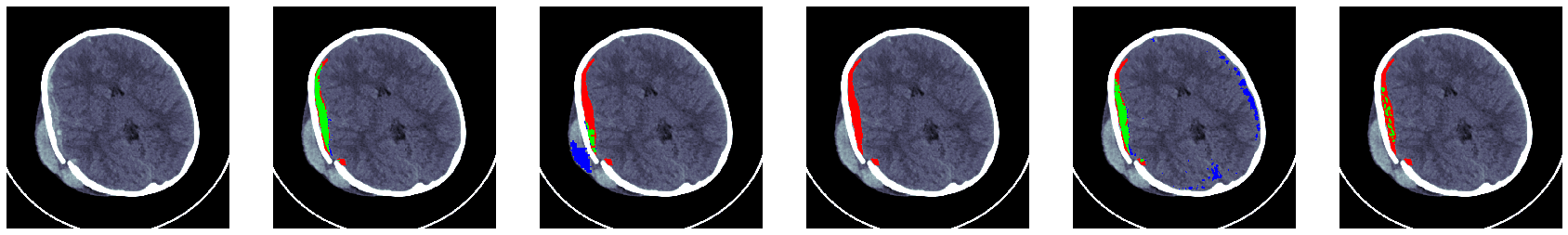}
         \caption{}
         \label{fig:visual3}
     \end{subfigure}
     \par\medskip
     \begin{subfigure}[b]{1.0\textwidth}
         \centering
         \includegraphics[width=\textwidth]{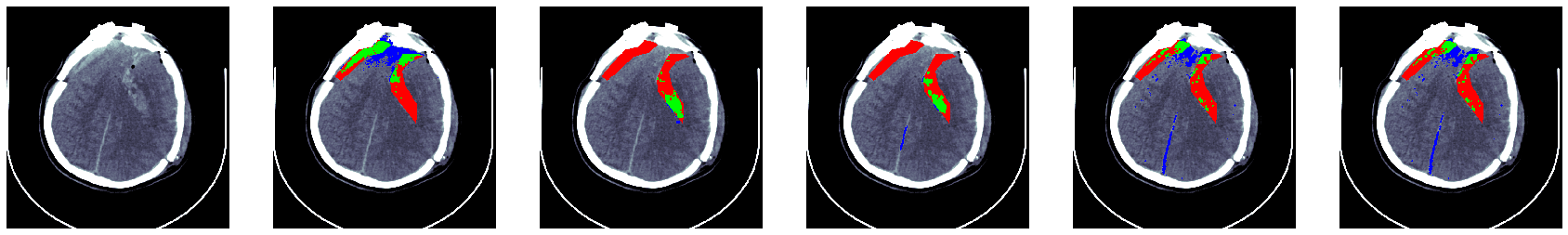}
         \caption{}
         \label{fig:visual4}
     \end{subfigure}
     \par\medskip
     
     \begin{subfigure}[b]{1.0\textwidth}
         \centering
         \includegraphics[width=\textwidth]{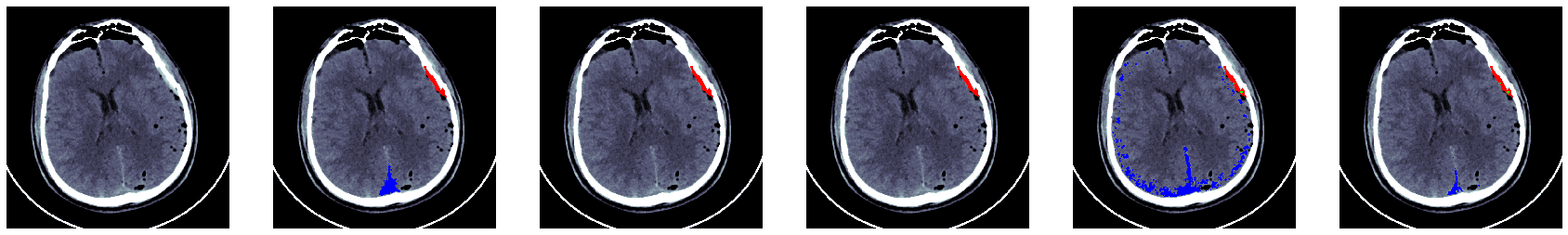}
         \caption{}
         \label{fig:visual6}
     \end{subfigure}
     \par\medskip
    \caption{Visual evaluation of the models. From left to right: 1) Original image, 2) Our model, 3) DeepBleed, 4) PItcHPERfeCT, 5) FCM 40, 6) FCM 45. Green indicates correctly segmented voxels, red indicates false negative voxels, and blue indicates false positive voxels.}
    \label{fig:visualanalysis}
\end{figure*}

Next, we perform visual analysis on the results to better understand where each model succeeds or fails. For the visualizations, we use the smaller PhysioNet dataset which provides ground truth segmentations so we can see precisely how each model's results differ from a radiologist's segmentation. We have hand-selected a set of CT scans that contains both successful and failed segmentations. The original and segmented images can be seen in Figure~\ref{fig:visualanalysis}. It should be noted that all of the chosen algorithms perform their detection on the full three-dimensional CT scan even though we only show one slice from each scan.

First, Image~\ref{fig:visual1} shows a wide subdural hemorrhage region on the right side along the skull. Our model matches the ground truth very accurately and only misses a small region at the top part of the hemorrhage as well as narrow slices along the edges. In the top part of the image, the hemorrhage's intensity becomes very similar to other brain tissue's intensity which makes it challenging to detect correctly. DeepBleed fails to detect most of the hemorrhage while PItcHPERFeCT only detects the highest-intensity regions of the hemorrhage leaving gaps at the locations that have a lower intensity. FCM40 detects most of the hemorrhage but includes large noisy regions along the edges of the brain and FCM45 misses the low-intensity regions of the hemorrhage.

Image~\ref{fig:visual2} contains epidural hemorrhage on the lower-left part of the image. Again, our model matches the ground truth nearly perfectly while only missing a narrow slice along the right edge of the hemorrhage. DeepBleed and PItcHPERFeCT both leave out the top part of the hemorrhage, most likely due to the slightly lower intensity within that area. FCM40 misses the areas along the top-right edge of the hemorrhage while marking small regions along the edges of the brain as hemorrhage. FCM45 does not detect any incorrect regions but misses large parts of the hemorrhage due to the lower intensity.

Image~\ref{fig:visual3} shows a subdural hemorrhage close to a skull fracture. Our algorithm detects most of the hemorrhage but misses a small, detached region below the fractured area. DeepBleed and PItcHPERFeCT miss most of the hemorrhage presumably because it is a very narrow region and has only a slightly higher intensity than the healthy brain tissue. DeepBleed also marks part of the edema located outside the skull as hemorrhage which indicates that the brain extraction did not work correctly with a fractured skull. Again, FCM40 misses part of the hemorrhage while including noise and FCM45 misses most of the hemorrhage.

Image~\ref{fig:visual4} demonstrates a more complicated situation where the skull is fractured and significant parts of the hemorrhage have a somewhat similar intensity to the other brain tissue. As the top-most part of the hemorrhage is close to higher-intensity brain tissue, our model mistakenly marks some of the healthy tissue as hemorrhage. At the same time, our model does not detect the lower regions of the hemorrhage which have a lower intensity. DeepBleed and PItcHPERFeCT show the opposite behavior as they only detect the lower region while missing the top parts of the hemorrhage. Both FCM40 and FCM45 perform poorly because the hemorrhage's intensity is very close to other brain tissue's intensity. Note especially the midline which has a higher intensity than most of the actual hemorrhage and which was incorrectly detected by FCM40, FCM45, and PItcHPERFeCT

Finally, Image~\ref{fig:visual6} shows a situation where all the models fail. It contains a very narrow hemorrhage region along the right side of the skull, which is hard to distinguish from noise, as even healthy patients typically have a slightly higher intensity near the edges of the brain. Instead of detecting this region, our model incorrectly detects the lower region which has a high intensity. DeepBleed and PItcHPERFeCT do not detect anything in this image while FCM40 and FCM45 incorrectly mark other regions near the edges as hemorrhage.

\section{Discussion}


As was shown in the previous section, our model can achieve similar or better results than the existing models on most hemorrhage types. Supervised models are typically expected to outperform unsupervised models due to their capability of learning more complicated patterns, but to our surprise, there was only one hemorrhage type where DeepBleed performed better than our model. Even in that category, our model outperformed the remaining models. There are multiple possible explanations for why the supervised models did not perform better. For example, their training data could differ from the public datasets that we used for evaluating the models, but it cannot be verified, as the training data has not been released. These possible differences could include e.g. a different distribution of hemorrhage types, different kinds of images (e.g. more high-intensity hemorrhages), or different CT scanner models. These differences could lead to the models overfitting to certain types of images and not generalizing well to other datasets.

When comparing the bounding box detection rates, the supervised models were significantly worse when the size was small. Our model's results could be explained by the fact that many of the smaller bounding boxes tend to be near the edges of a larger hemorrhage region. Due to our formulation, these regions are easily distinguished from noise as they are detected as a part of the larger region. This same can also explain part of the differences in the detection rates of the low-intensity bounding boxes.

Our model was also able to perform better than the FCM models. This was the expected outcome, as the FCM models are focused on the intensity levels, while our model is able to take into account the voxel neighbourhood as well. As can be seen from the visual comparisons in the previous section, our approach was able to avoid most of the high-intensity noise while simultaneously including the low-intensity parts of the hemorrhage as long as it was connected to a larger hemorrhage region. In general, our algorithm shows a good ability to adapt to hemorrhages with different intensities and shapes due to how the hemorrhage distributions are represented in our model.

Some of the most significant remaining challenges are 1) very small hemorrhage regions, 2) nearly-isodense hemorrhages, and 3) high-intensity healthy regions. The very small hemorrhages are not detected due to our algorithm's minimum threshold for hemorrhage sizes, which is is necessary to avoid including noise. Detecting some of the smaller hemorrhages could be possible by requiring a larger intensity if the hemorrhage is small, but even this approach would be missing some hemorrhages which are small and have a low intensity. Detecting nearly-isodense hemorrhages and ignoring high-intensity healthy regions are closely related problems. In both cases, hemorrhage and healthy tissue look very similar, and correct detection might require additional information. For example, it could be possible to take into account that certain regions of the image are likely to have higher intensities in general, e.g. the edges and the midline. The algorithm could have different intensity requirements for these regions, but it requires further analysis of the image to detect e.g. the midline correctly. For simplicity, we have not included this in the proposed algorithm.

\section{Summary}

Automated techniques for detecting intracranial hemorrhage can save significant amount of time when analysing CT scans, because a radiologist has to analyse dozens of images per patient. While many supervised techniques have been proposed for this problem, the lack of publicly-available training data makes them inaccessible to most. In this paper, we have proposed an unsupervised algorithm that is based on the mixture models and which can adaptively choose the appropriate number of clusters so that the hemorrhage is represented accurately. We have provided a comparison of our algorithm and a number of earlier algorithms which shows that our results are consistently better than the previously proposed algorithms in nearly all of the categories.

%
%
%
\bibliographystyle{IEEEtran}
\bibliography{mendeley,bibliography}

\end{document}